\documentclass[11pt,a4paper,hyphens]{article}
\usepackage{acl}
\usepackage{times}
\usepackage{latexsym}
\usepackage[T1]{fontenc}
\usepackage[utf8]{inputenc}

\usepackage[multiple]{footmisc}

\usepackage{url}

\usepackage{textcomp}
\usepackage{graphicx} 
\graphicspath{ {images/} }
\usepackage{comment}
\usepackage{xspace}
\usepackage{pifont}
\usepackage{amsmath}
\usepackage{txfonts}
\usepackage{scalefnt}
\usepackage{verbatim}

\newcommand{\textmc}[1]{\textsc{\scalefont{1.25}#1}}

\newcommand{\bert}{\textmc{bert}\xspace}
\newcommand{\bilstm}{\textmc{bilstm}\xspace}
\newcommand{\cnn}{\textmc{cnn}\xspace}
\newcommand{\crf}{\textmc{crf}\xspace}
\newcommand{\edgar}{\textmc{edgar}\xspace}
\newcommand{\edgarCorpus}{\textmc{edgar-corpus}\xspace}
\newcommand{\eu}{\textmc{eu}\xspace}
\newcommand{\lr}{\textmc{lr}\xspace}
\newcommand{\lstm}{\textmc{lstm}\xspace}

\newcommand{\finbert}{\textmc{fin-bert}\xspace}
\newcommand{\finerdata}{\textmc{f}i\textmc{ner}-139\xspace}
\newcommand{\genbert}{\textmc{g}en\textsc{bert}\xspace}
\newcommand{\ner}{\textmc{ner}\xspace}

\newcommand{\nltk}{\textmc{nltk}\xspace}
\newcommand{\macrof}{$\mathrm{m}$-$\mathrm{F_1}$\xspace}
\newcommand{\magn}{\textmc{{\small [}shape{\small ]}}\xspace}
\newcommand{\microf}{$\mathrm{\muup}$-$\mathrm{F_1}$\xspace}
\newcommand{\microp}{$\mathrm{\muup}$-$\mathrm{P}$\xspace}
\newcommand{\micror}{$\mathrm{\muup}$-$\mathrm{R}$\xspace}
\newcommand{\num}{\textmc{{\small [}num{\small ]}}\xspace}
\newcommand{\numbert}{\textmc{n}um\textsc{bert}\xspace}

\newcommand{\nlp}{\textmc{nlp}\xspace}
\newcommand{\oov}{\textmc{oov}\xspace}
\newcommand{\secbert}{\textmc{sec-bert}\xspace}
\newcommand{\secbertmag}{\textmc{sec-bert-shape}\xspace}
\newcommand{\secbertnum}{\textmc{sec-bert-num}\xspace}
\newcommand{\secbertfirst}{\textmc{sec-bert-first}\xspace}

\newcommand{\bilstmmag}{\textmc{bilstm-shape}\xspace}
\newcommand{\bilstmnum}{\textmc{bilstm-num}\xspace}
\newcommand{\spacy}{\textmc{spa}C\textmc{y}\xspace}
\newcommand{\uk}{\textmc{uk}\xspace}
\newcommand{\us}{\textmc{us}\xspace}
\newcommand{\ussec}{\textmc{sec}\xspace}
\newcommand{\wordvec}{\textmc{word}2\textmc{vec}\xspace}
\newcommand{\xbrl}{\textmc{xbrl}\xspace}
\newcommand{\xml}{\textmc{xml}\xspace}
\newcommand{\stdsign}{$\pm$\xspace}
\newcommand{\gaap}{\textmc{gaap}\xspace}

\usepackage{microtype}

\title{FiNER: Financial 
Numeric 
Entity Recognition for XBRL Tagging}

\author{
Lefteris Loukas$^{1,2}$, Manos Fergadiotis$^{1,2}$, Ilias Chalkidis$^{3}$, Eirini Spyropoulou$^{1}$, \\\textbf{Prodromos Malakasiotis}$^{1, 2}$, \textbf{Ion Androutsopoulos}$^{1,2}$, \textbf{George Paliouras}$^{1} $

\\ 
$^{1}$Institute of Informatics and Telecommunications, NCSR ``Demokritos''\\
$^{2}$Department of Informatics, Athens University of Economics and Business \\ 
$^{3}$Department of Computer Science, University of Copenhagen

\\

}
\newcommand\blfootnote[1]{
  \begingroup
  \renewcommand\thefootnote{}\footnote{#1}
  \addtocounter{footnote}{-1}
  \endgroup
}

\begin{document}
\maketitle
\begin{abstract}
Publicly traded companies are required to submit periodic reports with eXtensive Business Reporting Language (\xbrl) word-level tags. Manually tagging the reports is tedious and costly. We, therefore, introduce \xbrl tagging as a new entity extraction task for the financial domain and release \finerdata, a dataset of 1.1M sentences with gold \xbrl tags. Unlike typical entity extraction datasets, \finerdata uses a much larger label set of 139 entity types. Most annotated tokens are numeric, with the correct tag per token depending mostly on context, rather than the token itself. We show that subword fragmentation of numeric expressions harms \bert's performance, allowing word-level \bilstm{s} to perform better. To improve \bert's performance, we propose two simple and effective solutions that replace numeric expressions with pseudo-tokens reflecting original token shapes and numeric magnitudes. We also experiment with \finbert, an existing \bert model for the financial domain, and release our own \bert (\secbert), pre-trained on financial filings, which performs best. Through data and error analysis, we finally identify possible limitations to inspire future work on \xbrl tagging.
\end{abstract}

\section{Introduction}
\label{introduction} \label{sec:introduction}

\blfootnote{Source code: \url{https://github.com/nlpaueb/finer}}
\blfootnote{Correspondence: {\tt eleftheriosloukas@aueb.gr}}

Natural language processing (\nlp) for finance is an emerging research area \citep{econlp-2019, finnlp-2020, fnp-2020-joint}. Financial data are mostly reported in tables,but substantial information can also be found in textual form, e.g., in company filings, analyst reports, and economic news. Such information is useful in numerous financial intelligence tasks, like stock market prediction \citep{chen-etal-2019-group,yang-etal-2019-leveraging-finnlp1}, financial sentiment analysis \citep{Malo2014GoodDO-financial-sa-3, wang-etal-2013-financial-sa-1, akhtar-etal-2017-multilayer-financial-sa-2}, economic event detection \citep{ein-dor-etal-2019-financial-event-1,jacobs-etal-2018-economic-event-2, zhai-zhang-2019-forecasting-event-3}, and causality analysis \citep{tabari-etal-2018-causality-1, izumi-sakaji-2019-economic-causality-2}. 
In this work, we study how financial reports can be automatically enriched with word-level tags from the eXtensive Business Reporting Language (\xbrl), a tedious and costly task not considered so far.\footnote{See  \url{https://www.xbrl.org/the-standard/what/an-introduction-to-xbrl/} for an introduction to \xbrl.}  

\begin{figure}[t!] 
\centering

\fbox{
\includegraphics[width=0.9\columnwidth]{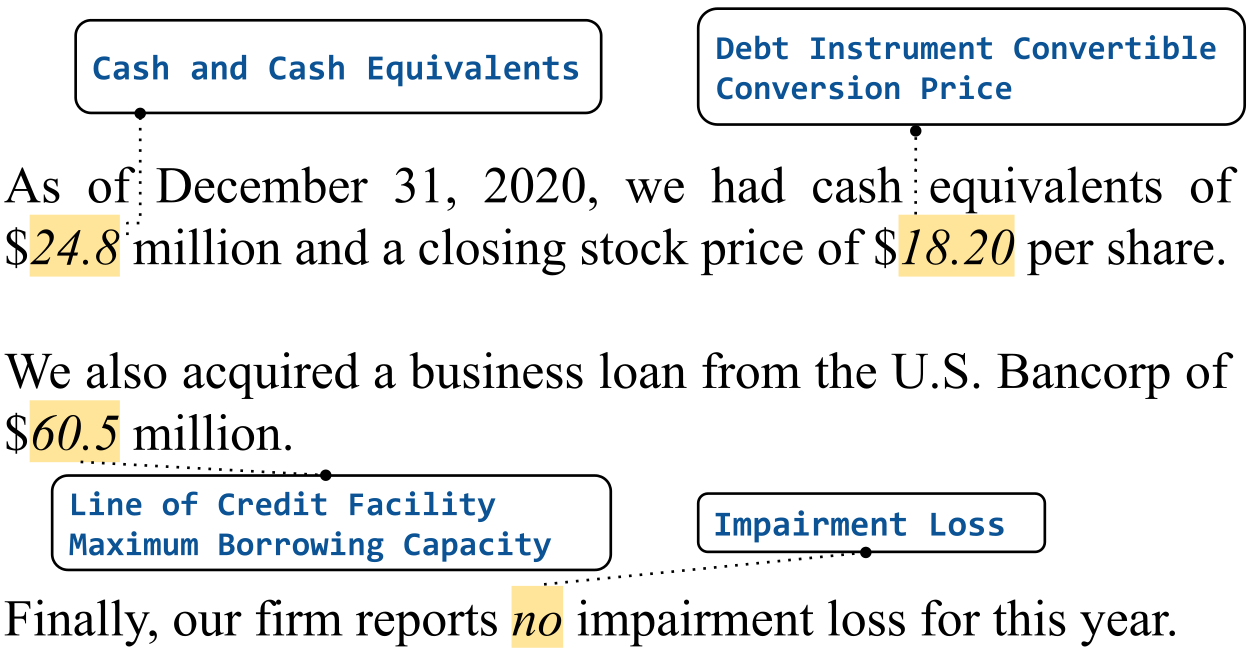}
}
\vspace*{-3mm}
\caption{Sentences from \finerdata, with \xbrl tags on numeric and non-numeric tokens. \xbrl tags are actually \xml-based and most tagged tokens are numeric.}
\label{fig:sample_sentences}
\vspace*{-2.5mm}
\end{figure}

To promote transparency among shareholders and potential investors, publicly traded companies are required to file periodic financial reports. These comprise multiple sections, including financial tables and text paragraphs, called \emph{text notes}. In addition, legislation in the \us, the \uk, the \eu and elsewhere requires the reports to be annotated with tags of \xbrl, an \xml-based language, to facilitate the processing of financial information.
The annotation of tables can be easily achieved by using company-specific pre-tagged table templates, since the structure and contents of the tables in the reports of a particular company rarely change. On the other hand, the unstructured and dynamic nature of text notes (Figure~\ref{fig:sample_sentences}) makes adding \xbrl tags to them much more difficult. Hence, we focus on automatically tagging text notes. 
Tackling this task could facilitate the annotation of new and old reports (which may not include \xbrl tags), e.g., by inspecting automatically suggested tags.

\begin{table}[t]
\large
\resizebox{\columnwidth}{!}
{
\centering

\begin{tabular}{l|c|c}
Dataset                                   & Domain          & Entity Types \\\hline 
\textmc{conll}-2003                       & Generic         & 4   \\
\textmc{ontonotes-v}5                     & Generic         & 18  \\
\textmc{ace}-2005                         & Generic         & 7 \\
\textmc{genia}                            & Biomedical      & 36 \\
\citet{chalkidis2019neural}               & Legal           & 14\\
\citet{francis2019transfer-related-2019}  & Financial       & 9 \\
\finerdata (ours)                         & Financial       & \textbf{139} \\
\hline
\end{tabular}
}
\vspace{-2.5mm}
\caption{Examples of 
previous entity extraction datasets. Information about the first four  from \citet{conll2003, ontonotesv5, ace2005, genia2003}.}
\label{tab:ner-stats}
\vspace*{-2.5mm}
\end{table}

Towards this direction, we release \finerdata, a new dataset of 1.1M sentences with gold \xbrl tags, from annual and quarterly reports of publicly traded companies obtained from the \us Securities and Exchange Commission (\ussec).
Unlike other entity extraction tasks, like named entity recognition (\ner) or contract element extraction (Table~\ref{tab:ner-stats}), which typically require identifying entities of a small set of common types (e.g., persons, organizations), \xbrl defines approx.\ 6k entity types. As a first step, we consider the 139 most frequent \xbrl entity types,
still a much larger label set than usual.

Another important difference from typical 
entity extraction is that most tagged tokens ($\sim$91\%) in the text notes we consider are numeric, with the correct tag per token depending mostly on context, not the token itself (Figure~\ref{fig:sample_sentences}). The abundance of numeric tokens also leads to a very high ratio of out-of-vocabulary (\oov) tokens, approx.\ 10.4\% when using a custom \wordvec \cite{word-embeddings} model trained on our corpus. When using subwords, e.g., in models like \bert \cite{bert}, there are no \oov tokens, but numeric expressions get excessively fragmented, making it difficult for the model to gather information from the fragments and correctly tag them all. In our experiments, this is evident by the slightly better performance of stacked \bilstm{s} \citep{graves2013, lample-2016-neural} operating on word embeddings compared to \bert. The latter improves when using a \crf \cite{crf-1} layer, which helps avoid assigning nonsensical sequences of labels to the fragments (subwords) of numeric expressions. 

To further improve \bert's performance, we propose two simple and effective solutions that replace numeric expressions with pseudo-tokens reflecting the original token shapes and magnitudes. We also experiment with \finbert \citep{yang2020finbert}, an existing \bert model for the financial domain, and release our own family of \bert models, pre-trained on 200k financial filings, achieving the best overall performance.

Our key contributions are:\vspace*{-2mm} 

\urlstyle{rm} 

\begin{enumerate}

\item We introduce \xbrl tagging, a new financial \nlp task for a real-world need, and we release \finerdata, the first \xbrl tagging dataset.\footnote{\url{https://huggingface.co/datasets/nlpaueb/finer-139}}\vspace*{-2mm}

\item We provide extensive experiments using
\bilstm{s} and \bert with generic or in-domain pre-training, which establish strong baseline results for future work on \finerdata.\vspace*{-2mm}

\item We show that replacing numeric tokens with pseudo-tokens reflecting token shapes and magnitudes significantly boosts the performance of \bert-based models in this task.\vspace*{-2mm}

\item We release a new family of \bert models (\secbert, \secbertnum, \secbertmag) pre-trained on 200k financial filings that obtains the best results on \finerdata.\footnote{\url{https://huggingface.co/nlpaueb/sec-bert-base}}\footnote{\url{https://huggingface.co/nlpaueb/sec-bert-num}}\footnote{\url{https://huggingface.co/nlpaueb/sec-bert-shape}}

\end{enumerate}
\section{Related Work} \label{sec:related}

\noindent\textbf{Entity extraction:} 
\xbrl tagging differs from \ner and other previous entity extraction tasks (Table \ref{tab:ner-stats}), like  contract element extraction \cite{chalkidis2019neural}. Crucially, in \xbrl tagging there is a much larger set of entity types (6k in full \xbrl, 139 in \finerdata), most tagged tokens are numeric ($\sim$91\%), and the correct tag highly depends on context. 
In most \ner datasets, numeric expressions are classified in generic entity types like `amount' or `date' \citep{classic-ner-2}; this can often be achieved with regular expressions that look for common formats of numeric expressions, and the latter are often among the easiest entity types in \ner datasets. By contrast, although it is easy to figure out that the first three highlighted expressions of Figure~\ref{fig:sample_sentences} are amounts, assigning them the correct \xbrl tags requires carefully considering their context. Contract element extraction \cite{chalkidis2019neural} also requires considering the context of dates, amounts etc.\ to distinguish, for example, start dates from end dates, total amounts from other mentioned amounts, but the number of entity types in \finerdata is an order of magnitude larger (Table \ref{tab:ner-stats}) and the full tag set of \xbrl is even larger (6k).

\vspace{2mm}

\noindent\textbf{Financial \ner:} Previous financial \ner applications use at most 9 (generic) class labels. \citet{salinas-alvarado-etal-2015-domain-related-2015} investigated \ner in finance to recognize organizations, persons, locations, and miscellaneous entities on 8 manually annotated \ussec financial agreements using \crf{s}. \citet{francis2019transfer-related-2019} experimented with transfer learning by unfreezing different layers of a \bilstm with a \crf layer, pre-trained on invoices, to extract 9 entity types with distinct morphological patterns (e.g., \textmc{iban}, company name, date, total amount). Also, \citet{Hampton2015-related-3, Hampton2016-related-2} applied a Maximum Entropy classifier, \crf{s}, and handcrafted rules to London Stock Exchange filings to detect 9 generic entity types (e.g., person, organization, location, money, date, percentages). Finally, \citet{kumar-etal-2016-experiments-related-1-demo} extended the work of \citet{stanford-ner} and built a financial entity recognizer of dates, numeric values, economic terms in \ussec and non-\ussec documents, using numerous handcrafted text features. By contrast, \finerdata uses a specialized set of 139 highly technical economic tags derived from the real-world need of \xbrl tagging, and we employ no handcrafted features. 

\vspace{2mm}

\noindent\textbf{Numerical reasoning:} Neural numerical reasoning studies how to represent numbers to solve numeracy tasks, e.g., compare numbers, understand mathematical operations mentioned in a text etc. \citet{numbert-2020-zhang} released \numbert, a Transformer-based model that handles numerical reasoning tasks by representing numbers by their scientific notation and applying subword tokenization. On the other hand, \genbert \cite{genbert-geva-20} uses the decimal notation and digit-by-digit tokenization of numbers. Both models attempt to deal with the problem that word-level tokenization often turns numeric tokens to \textmc{oov}s \cite{thawani2021representing}. This is important, because numerical reasoning requires modeling the exact value of each numeric token. In \finerdata, the correct \xbrl tags of numeric tokens depend much more on their contexts and token shapes than on their exact numeric values (Fig.~\ref{fig:sample_sentences}). Hence, these methods are not directly relevant. \genbert{'s} digit-by-digit tokenization would also lead to excessive fragmentation, which we experimentally find to harm performance.

\section{Task and Dataset}
\label{sec:dataset-task}

Traditionally, business filings were simply rendered in plain text. Thus, analysts and researchers needed to manually identify, copy, and paste each amount of interest (e.g., from filings to spreadsheets). With \xbrl-tagged filings, identifying and extracting amounts of interest (e.g., to spreadsheets or databases) can be automated. More generally, \xbrl facilitates the machine processing of financial documents.
Hence, \xbrl-tagged financial reports are required in several countries, as already noted (Section~\ref{sec:introduction}). However, manually tagging reports with \xbrl tags is tedious and resource-intensive. Therefore, we release \finerdata to foster research towards automating \xbrl tagging. 

\begin{figure}[t] 
\centering
\includegraphics[width=\columnwidth]{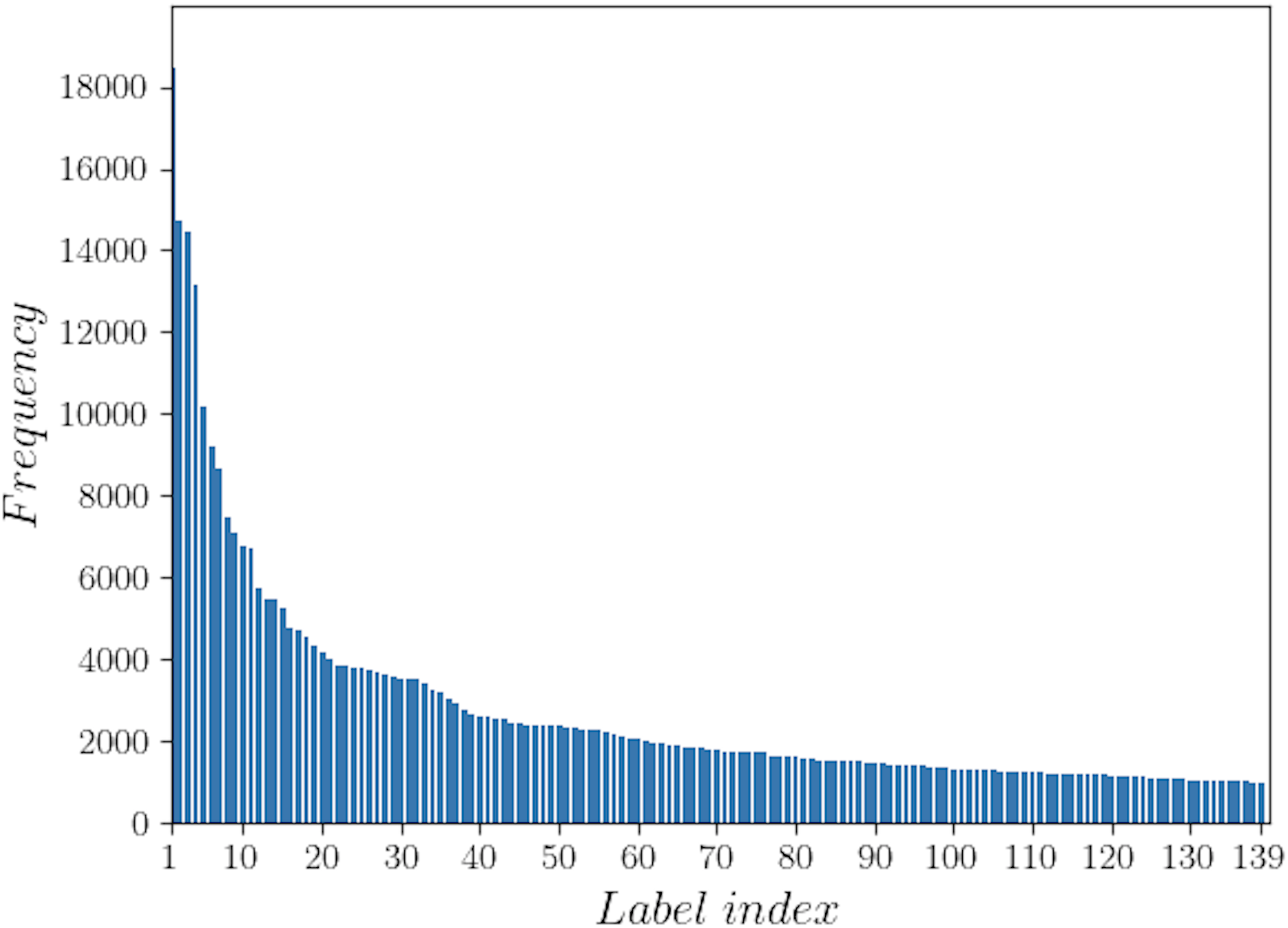}
\vspace*{-6mm}
\caption{Frequency distribution of the 139 \xbrl tags used in this work over the entire \finerdata dataset. Label indices shown instead of tag names to save space.
}
\label{fig:app_label-distribution}
\vspace*{-4.5mm}
\end{figure}

\finerdata was compiled from approx.\ 10k annual and quarterly English reports (filings) of publicly traded companies downloaded from \ussec's \edgar system.\footnote{\url{https://www.sec.gov/edgar/}} The downloaded reports span a 5-year period, from 2016 to 2020. They are annotated with \xbrl tags by professional auditors and describe the performance and projections of the companies. We used regular expressions to extract the text notes from the \textit{Financial Statements Item} of each filing, which is the primary source of \xbrl tags in annual and quarterly reports.

\xbrl taxonomies have many different attributes, making \xbrl tagging challenging even for humans \cite{Baldwin2006XBRLAI, Hoitash2018MeasuringAR}. Furthermore, each jurisdiction has its own \xbrl taxonomy. Since we work with \us documents, our labels come from \us-\gaap.\footnote{\url{www.xbrl.us/xbrl-taxonomy/2020-us-gaap/}} Since this is the first effort towards automatic \xbrl tagging, we chose to work with the most essential and informative attribute, the \textit{tag names}, which populate our label set. Also, since \xbrl tags change periodically, we selected the 139 (out of 6,008) most frequent \xbrl tags with at least 1,000 appearances in \finerdata. The distribution of these tags seems to follow a power law (Figure~\ref{fig:app_label-distribution}), hence most of the 6k \xbrl tags that we did not consider are very rare. We used the \textmc{iob}2 annotation scheme to distinguish tokens at the beginning, inside, or outside of tagged expressions, which leads to 279 possible token labels.

\begin{table}[t]
\LARGE
\centering
\resizebox{\columnwidth}{!}
{
\begin{tabular}{lccc}
\hline Subset & Sentences (S) & Avg.\ Tokens/S & Avg.\ Tags/S\\ \hline
Train                  & 900,384 & 44.7 \stdsign 33.9 & 1.8 \stdsign 1.2\\
Dev                    & 112,494 & 45.4 \stdsign 35.9 & 1.7 \stdsign 1.2\\
Test                   & 108,378 & 46.5 \stdsign 38.9 & 1.7 \stdsign 1.1\\
\hline
\end{tabular}
}
\vspace{-2.5mm}
\caption{\finerdata statistics, using \spacy's tokenizer and the 139 tags of this work ($\pm$ standard deviation). 
}
\label{tab:dataset-statistics}
\vspace*{-5mm}
\end{table}

We split the text notes into 1.8M sentences, the majority of which ($\sim$90\%) contained no tags.\footnote{We use \nltk \cite{nltk} for sentence splitting.} The sentences are also \textmc{html}-stripped, normalized, and lower-cased. To avoid conflating trivial and more difficult cases, we apply heuristic rules to discard sentences that can be easily flagged as almost certainly requiring no tagging; in a real-life setting, the heuristics, possibly further improved, would discard sentences that do not need to be processed by the tagger. The heuristic rules were created by inspecting the training subset and include regular expressions that look for amounts and other expressions that are typically annotated. Approx.\ 40\% of the 1.8M sentences were removed, discarding only 1\% of tagged ones. We split chronologically the remaining sentences into training, development, and test sets with an 80/10/10 ratio (Table~\ref{tab:dataset-statistics}).

\section{Baseline Models} \label{sec:baselines}

\noindent\textbf{\spacy} \citep{spacy} is an open-source \nlp library.\footnote{We used \spacy v.2.3; see \url{https://spacy.io/}.} It includes an industrial \ner that uses  word-level Bloom embeddings \citep{bloom-embeddings} and residual Convolutional Neural Networks (\cnn{s}) \citep{residual-cnns}. We trained \spacy's \ner from scratch on \finerdata.
\vspace{2mm}

\noindent\textbf{\bilstm:}
This baseline uses a stacked bidirectional Long-Short Term Memory (\lstm) network \citep{graves2013, lample-2016-neural} with residual connections. Each token $t_i$ of a sentence $S$ is mapped to an embedding and passed through the \bilstm stack to extract the corresponding contextualized embedding. A shared multinomial logistic regression (\lr) layer operates on top of each contextualized embedding to predict the correct label.
We use the \wordvec embeddings \citep{word-embeddings, word-embeddings-2} of \citet{loukas-etal-2021-edgar}.\vspace{2mm}

\noindent\textbf{\bert:} This is similar to \bilstm, but now we fine-tune \textmc{bert-base} \citep{bert} to extract contextualized embeddings of subwords. Again, a multinomial \lr layer operates on top of the contextualized embeddings to predict the correct label of the corresponding subword. 
\vspace{2mm}

\noindent\textbf{\crf{s}:} In this case, we replace the \lr layer of the previous two models with a Conditional Random Field (\crf) layer \citep{crf-1}, which has been shown to be beneficial in several token labeling tasks \citep{crf-ner-1,lample-2016-neural, chalkidis-etal-2020-legal}.\footnote{We use a linear-chain \crf layer with log-likelihood optimization and Viterbi decoding.}
\section{Baseline Results}\label{sec:baseline_results}

We report micro-$\mathrm{F_1}$ (\microf) and macro-$\mathrm{F_1}$ (\macrof) at the entity level, i.e., if a gold tag annotates a multi-word span, a model gets credit only if it tags the exact same span. This allows comparing more easily methods that label words vs.\ subwords.

Table~\ref{tab:baseline-results} shows that \spacy performs poorly, possibly due to the differences from typical token labeling tasks, i.e., the large amount of entity types, the abundance of numeric tokens, and the fact that in \finerdata the tagging decisions depend mostly on context. 
Interestingly enough, \bilstm (with word embeddings) performs slightly better than \bert. However, when a \crf layer is added, \bert achieves the best results, while the performance of \bilstm (with word embeddings) deteriorates significantly, contradicting previous studies. 

\begin{table}[h]
\large
\centering
\resizebox{\columnwidth}{!}
{
\begin{tabular}{ l|cc}
 \hline
 Baseline methods & \microf & \macrof \\
 \hline\hline
 \spacy (words)         & 48.6 $\pm$ 0.4             & 37.6 $\pm$ 0.2\\
 \bilstm (words) & \underline{77.3} $\pm$ 0.6 & \underline{73.8} $\pm$ 1.8\\
 \bilstm (subwords) & 71.3 \stdsign 0.2 & 68.6 \stdsign 0.2 \\
 \bert (subwords)         & 75.1 $\pm$ 1.1             & 72.6 $\pm$ 1.4\\\hline
 \bilstm (words) + \crf  & 69.4 $\pm$ 1.2             & 67.3 $\pm$ 1.6\\
 \bilstm (subwords) + \crf  & 76.2 \stdsign 0.2 & 73.4 \stdsign 0.3\\
 \bert (subwords) + \crf    & \textbf{78.0} $\pm$ 0.5    & \textbf{75.2} $\pm$ 0.6\\\hline
 
\end{tabular}
}
\vspace{-2.5mm}
\caption{Entity-level \microf and \macrof (\%, avg.\ of 3 runs with different random seeds, $\pm$ std.\ dev.) on test data.}

\label{tab:baseline-results}
\vspace*{-2.5mm}
\end{table}
We hypothesize that the inconsistent effect of \crf{s} 
is due to tokenization differences. When using \bert's subword tokenizer, there are more decisions that need to be all correct for a tagged span to be correct (one decision per subword) than when using word tokenization (one decision per word). Thus, it becomes more difficult for subword models to avoid nonsensical sequences of token labels, e.g., labeling two consecutive subwords as beginning and inside of different entity types, especially given the large set of 279 labels (Table~\ref{tab:ner-stats}). The \crf layer on top of subword models helps reduce the nonsensical sequences of labels.

On the other side, when using words as tokens, there are fewer opportunities for nonsensical label sequences, because there are fewer tokens. For instance, the average number of subwords and words per gold span is 2.53 and 1.04, respectively. Hence, it is easier for the \bilstm to avoid predicting nonsensical sequences of labels and the \crf layer on top of the \bilstm (with word embeddings) has less room to contribute and mainly introduces noise (e.g., it often assigns low probabilities to acceptable, but less frequent label sequences).
With the \crf layer, the model tries to maximize both the confidence of the \bilstm for the predicted label of each word and the probability that the predicted sequence of labels is frequent. When the \bilstm on its own rarely predicts nonsensical sequences of labels, adding the \crf layer rewards commonly seen sequences of labels, even if they are not the correct labels, without reducing the already rare nonsensical sequences of labels.

To further support our hypothesis, we repeated the \bilstm experiments, but with subword (instead of word) embeddings, trained on the same vocabulary with \bert. Without the \crf, the subword \bilstm performs much worse than the word \bilstm (6 p.p drop in \microf), because of the many more decisions and opportunities to predict nonsensical label sequences. The \crf layer substantially improves the performance of the subword \bilstm (4.9 p.p.\ increase in \microf), as expected, though the word \bilstm (without \crf) is still better, because of the fewer opportunities for nonsensical predictions.
A drawback of \crf{s} is that they significantly slow down the models both during training and inference, especially when using large label sets \citep{crfs-fail-1}, as in our case. Hence, although \bert with \crf was the best model in Table~\ref{tab:baseline-results}, we wished to improve \bert's performance further without employing \crf{s}.

\section{Fragmentation in \bert}\label{sec:fragmentation_bert}
In \finerdata, the majority (91.2\%) of the gold tagged spans are numeric expressions, which cannot all be included in \bert's finite vocabulary; e.g., the token `{\small \texttt{9,323.0}}' is split into five subword units, {\small [`\texttt{9}', `\texttt{\#\#,}', `\texttt{\#\#323}', `\texttt{\#\#.}', `\texttt{\#\#0}']}, while the token `{\small \texttt{12.78}}' is split into {\small [`\texttt{12}', `\texttt{\#\#.}', `\texttt{\#\#78}']}. The excessive fragmentation of numeric expressions, when using subword tokenization, harms the performance of the subword-based models (Table \ref{tab:baseline-results}), because it increases the probability of producing nonsensical sequences of labels, as already discussed. We, therefore, propose two simple and effective solutions to avoid the over-fragmentation of numbers.\vspace{2mm}

\begin{figure}[t!] 
\centering

\fbox{
\includegraphics[width=0.95\columnwidth]{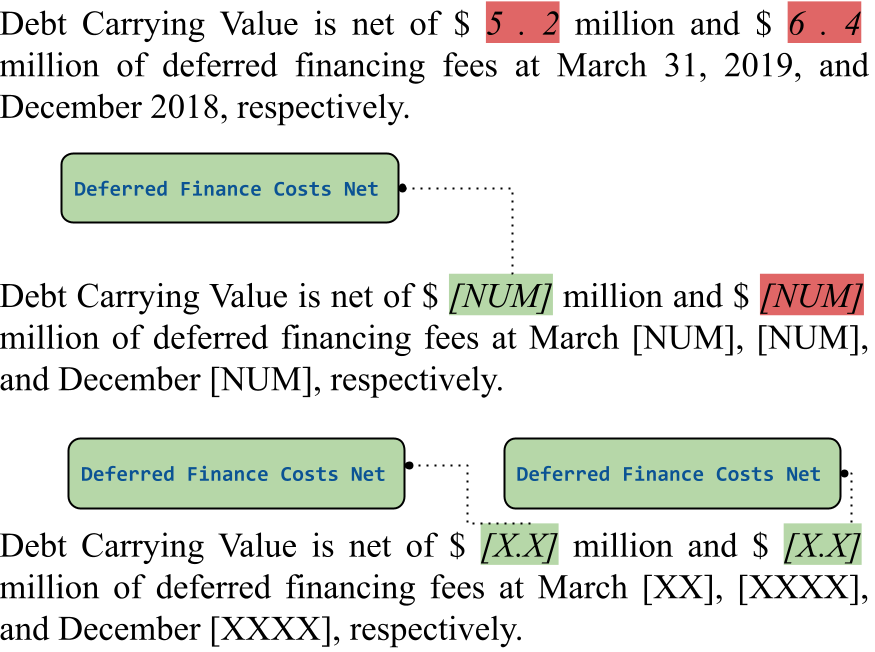}
}
\vspace*{-4mm}
\caption{\xbrl tag predictions of \bert (top), \bert + \num (middle), \bert + \magn (bottom) for the same sentence. \bert tags incorrectly the amounts in red. \bert + \num\ and \bert + \magn tag them more successfully (green indicates correct tags).}

\label{fig:different_shapes}
\vspace*{-3.5mm}
\end{figure}

\noindent\textbf{\bert + \num:} 
We detect numbers using regular expressions and replace each one with a single \num pseudo-token, which cannot be split. The pseudo-token is added to the \bert vocabulary, and its representation is learned during fine-tuning. This allows handling all numeric expressions in a uniform manner, disallowing their fragmentation.\vspace{2mm}

\noindent\textbf{\bert + \magn:} We replace numbers with pseudo-tokens that cannot be split and represent the number's shape. For instance, `{\small \texttt{53.2}}' becomes `{\small $\mathrm{[XX.X]}$}', and `{\small \texttt{40,200.5}}' becomes `{\small $\mathrm{[XX}$,$\mathrm{XXX.X]}$}'. We use 214 special tokens that cover all the number shapes of the training set. Again, the representations of the pseudo-tokens are fine-tuned, and numeric expressions (of known shapes) are no longer fragmented. The shape pseudo-tokens also capture information about each number's magnitude; the intuition is that numeric tokens of similar magnitudes may require similar \xbrl tags. Figure \ref{fig:different_shapes} illustrates the use of \num and \magn pseudo-tokens.

\begin{table*}[t]
\centering
\resizebox{\textwidth}{!}
{
\begin{tabular}{ l|ccc|ccc}
 \hline
 &
 \multicolumn{3}{c}{\textmc{development}} &  \multicolumn{3}{c}{\textmc{test}} \\
 & \microp & \micror & \microf   & \microp & \micror & \microf \\
 \hline
 \hline

 \bert       
 &  74.9 $\pm$ 1.5
& 82.0 $\pm$ 1.3
& 78.2 $\pm$ 1.4 
& 71.5 $\pm$ 1.1
& 79.6 $\pm$ 1.4
& 75.1 $\pm$ 1.1 \\

 \bert{ + \crf}       
 &\underline{78.3} $\pm$ 0.8
 &\underline{83.6} $\pm$ 0.4
 &\underline{80.9} $\pm$ 0.3
 
 &\underline{75.0} $\pm$ 0.9
 &\underline{81.2} $\pm$ 0.2
 &\underline{78.0} $\pm$ 0.5
 \\

 \hline
 \bert + \num
 & 79.4 $\pm$ 0.8
 & \underline{83.0} $\pm$ 0.9
 & 81.2 $\pm$ 0.9
 & 76.0 $\pm$ 0.6
 & \underline{80.7} $\pm$ 0.8
 & 78.3 $\pm$ 0.7 
 \\
 
 \bert + \magn
 & \underline{82.1} $\pm$ 0.6
 & 82.6 $\pm$ 0.4
 & \underline{82.3} $\pm$ 0.2
 & \underline{78.7} $\pm$ 0.5
 & 80.1 $\pm$ 0.2
 & \underline{79.4} $\pm$ 0.2
 \\
 \hline
 \finbert                       
 & 73.9 $\pm$ 1.3
 & 81.4 $\pm$ 0.7
 & 77.5 $\pm$ 1.0
 & 70.2 $\pm$ 1.2
 & 78.7 $\pm$ 0.7
 & 74.0 $\pm$ 1.1
 \\
 \finbert + \num
 & 81.1 $\pm$ 0.1
 & 82.5 $\pm$ 1.2
 & 81.8 $\pm$ 0.1
 & 77.9 $\pm$ 0.1
 & 79.9 $\pm$ 0.7
 & 78.8 $\pm$ 0.3
 \\
 \finbert + \magn

 & \underline{82.3} $\pm$ 1.7
 & \underline{84.0} $\pm$ 1.2
 & \underline{83.2} $\pm$ 1.4
 & \underline{79.0} $\pm$ 1.6
 & \underline{81.2} $\pm$ 1.1
 & \underline{80.1} $\pm$ 1.4
 \\
 \hline
 \secbert (ours) 
 
 & 75.2 $\pm$ 0.4
 & 82.7 $\pm$ 0.5
 & 78.8 $\pm$ 0.1
 & 71.6 $\pm$ 0.4
 & 80.3 $\pm$ 0.5
 & 75.7 $\pm$ 0.1 
 \\

 \secbertnum (ours)                
 & 82.5 $\pm$ 2.1
 & 84.4 $\pm$ 1.2
 & 83.7 $\pm$ 1.7
 & 79.0 $\pm$ 1.9
 & 82.0 $\pm$ 0.9
 & 80.4 $\pm$ 1.4
 \\
 \secbertmag (ours)              
 & \textbf{84.8} $\pm$ 0.2
 & \textbf{85.8} $\pm$ 0.2
 & \textbf{85.3} $\pm$ 0.0
 & \textbf{81.0} $\pm$ 0.2
 & \textbf{83.2} $\pm$ 0.1
 & \textbf{82.1} $\pm$ 0.1
 \\
 
\end{tabular}
}
\vspace*{-1.5mm}
\caption{{Entity-level micro-averaged $\mathrm{P}$, $\mathrm{R}$, $\mathrm{F_1}$ $\pm$ std.\ dev.\ (3 runs) on the dev.\ and test data for \bert{-based} models.}}
\label{tab:variants-results}
\end{table*}

\section{In-domain Pre-training}\label{sec:in_domain_knowledge}
Driven by the recent findings that pre-training language models on specialized domains is beneficial for downstream tasks \citep{alsentzer2019clinical, beltagy-etal-2019-scibert, yang2020finbert,chalkidis-etal-2020-legal}, we explore this direction in our task which is derived from the financial domain.\vspace{2mm}

\noindent\textbf{\finbert:} We fine-tune \finbert \citep{yang2020finbert}, which is pre-trained on a financial corpus from \ussec documents, earnings call transcripts, and analyst reports.\footnote{We use the \textmc{finbert-finvocab-uncased} version from \url{https://github.com/yya518/FinBERT}.} 
The 30k subwords vocabulary of \finbert is built from scratch from its pre-training corpus.
Again, we utilize \finbert with and without our numeric pseudo-tokens, whose representations are learned during fine-tuning.\vspace{2mm}

\noindent\textbf{\secbert:} We also release our own family of \bert models. Following the original setup of \citet{bert}, we pre-trained \bert from scratch on \edgarCorpus, a collection of financial documents released by \citet{loukas-etal-2021-edgar}. The resulting model, called \secbert, has a newly created vocabulary of 30k subwords. To further examine the impact of the proposed \num and \magn special tokens, we also pre-trained two additional \bert variants, \secbertnum and \secbertmag, on the same corpus, having replaced all numbers by \num or \magn pseudo-tokens, respectively. 
In this case, the representations of the pseudo-tokens are learned during pre-training and they are updated during fine-tuning. 

\section{Improved \bert Results}
Table~\ref{tab:variants-results} reports micro-averaged precision, recall, and $\mathrm{F_1}$ on development and test data. As with Table ~\ref{tab:baseline-results}, a \lr layer is used on top of each embedding to predict the correct label, unless specified otherwise.

Focusing on the second zone, we observe that the \num pseudo-token improves \bert's results, as expected, since it does not allow numeric expressions to be fragmented. The results of \bert + \num are now comparable to those of \bert + \crf. Performance improves further when utilizing the shape pseudo-tokens (\bert + \magn), yielding 79.4 \microf and showing that information about each number's magnitude is valuable in \xbrl tagging.

Interestingly, \finbert (3rd zone) performs worse than \bert despite its pre-training on financial data. Similarly to \bert, this can be attributed to the fragmentation of numbers (2.5 subwords per gold tag span). Again, the proposed pseudo-tokens (\num, \magn) alleviate this problem and allow \finbert to leverage its in-domain pre-training in order to finally surpass the corresponding \bert variants, achieving an 80.1 \microf test score.

Our new model, \secbert (last zone), which is pre-trained on \ussec reports, performs better than the existing \bert and \finbert models, when no numeric pseudo-tokens are used. However, \secbert is still worse than  \bert with numeric pseudo-tokens (75.7 vs.\ 78.3 and 79.4 test \microf), suffering from number fragmentation (2.4 subwords per gold tag span). \secbert (without pseudo-tokens) also performs worse than the \bilstm with word embeddings (75.7 vs.\ 77.3 \microf, cf.\ Table~\ref{tab:baseline-results}). However, when the proposed pseudo-tokens are used, \secbertnum and \secbertmag achieve the best overall performance, boosting the test \microf to 80.4 and 82.1, respectively. This indicates that learning to handle numeric expressions during model pre-training is a better strategy than trying to acquire this knowledge only during fine-tuning.

\section{Additional Experiments}

\subsection{Subword pooling}\label{sec:subword_pooling}
An alternative way to bypass word fragmentation is to use subword pooling for each word. \citet{acs-etal-2021-subword} found that for \ner tasks, it is better to use the \emph{first} subword only, i.e., predict the label of an entire word from the contextualized embedding of its first subword only; they compared to several other methods, such as using only the \emph{last} subword of each word, or combining the contextualized embeddings of all subwords with a self-attention mechanism. Given this finding, we conducted an ablation study and compare (i) our best model (\secbert) with \emph{first} subword pooling (denoted \secbertfirst) to (ii) \secbert with our  special tokens (\secbertnum, \secbertmag), which avoid segmenting numeric tokens.

Table \ref{tab:subword_pooling} shows that, in \xbrl tagging, using the proposed special tokens is comparable (\secbertnum) or better (\secbertmag) than performing \emph{first} pooling (\secbertfirst). It might be worth trying other pooling strategies as well, like \textit{last}-pooling or subword self-attention pooling. It's worth noting, however, that the latter will increase the training and inference times.

\begin{table}[h]
\small
\centering
\resizebox{0.95\columnwidth}{!}
{
\begin{tabular}{ l|cc}
 \hline
   & \microf & \macrof \\
 \hline\hline
\secbert & 78.8 \stdsign 0.1 & 72.6 \stdsign 0.4 \\
\hline \secbertfirst & 79.9 \stdsign 1.2 & 77.1 \stdsign 1.7 \\
\secbertnum & 80.4 \stdsign 1.4 & 78.9 \stdsign 1.3\\
 \secbertmag & \textbf{82.1} \stdsign 0.1 & \textbf{80.1} \stdsign 0.2 \\
 \hline
 
\end{tabular}
}
\caption{Entity-level \microf and \macrof (\%, avg.\ of 3 runs with different random seeds, $\pm$ std.\ dev.) on test data using different ways to alleviate fragmentation.}
\label{tab:subword_pooling}
\vspace*{-2.5mm}
\end{table}

\subsection{Subword \bilstm with \num and \magn} \label{ablation_study}
To further investigate the effectiveness of our pseudo-tokens, we incorporated them in the \bilstm operating on subword embeddings (3rd model of Table~\ref{tab:baseline-results}). Again, we replace each number by a single \num pseudo-token or one of 214 \magn pseudo-tokens, for the two approaches, respectively. These replacements also happen when pre-training \wordvec subword embeddings; hence, an embedding is obtained for each pseudo-token.

Table~\ref{tab:ablation-study} shows that \bilstmnum outperforms the \bilstm subword model. \bilstmmag further improves performance and is the best \bilstm subword model overall, surpassing the subword \bilstm with \crf, which was the best subword \bilstm model in Table~\ref{tab:baseline-results}. These results further support our hypothesis that the \num and \magn pseudo-tokens help subword models successfully generalize over numeric expressions, with \magn being the best of the two approaches, while also avoiding the over-fragmentation of numbers.

\begin{table}[h]
\LARGE
\centering
\resizebox{\columnwidth}{!}
{
\begin{tabular}{ l|cc}
 \hline
   & \microf & \macrof \\
 \hline\hline
\bilstm (subwords) & 71.3 \stdsign 0.2 & 68.6 \stdsign 0.2 \\
\bilstm (subwords) + \crf  & 76.2 \stdsign 0.2 & 73.4 \stdsign 0.3\\\hline
 \bilstmnum (subwords)  & 75.6 \stdsign 0.3 & 72.7 \stdsign 0.4 \\
 \bilstmmag (subwords) & \textbf{76.8} \stdsign 0.2 & \textbf{74.1} \stdsign 0.3 \\
 \hline
 
\end{tabular}
}
\caption{Entity-level \microf and \macrof (\%, avg.\ of 3 runs with different random seeds, $\pm$ std.\ dev.) on test data for \bilstm models with \num and \magn tokens.}

\label{tab:ablation-study}
\vspace*{-2.5mm}
\end{table}

\subsection{A Business Use Case}\label{business_study} \label{sec:business_study}
Since \xbrl tagging is derived from a real-world need, it is crucial to analyze the model's performance in a business use case. After consulting with experts of the financial domain, we concluded that one practical use case would be to use an \xbrl tagger as a recommendation engine that would propose the $k$ most probable \xbrl tags for a specific token selected by the user. The idea is that an expert (e.g., accountant, auditor) knows beforehand the token(s) that should be annotated and the tagger would assist by helping identify the appropriate tags more quickly. Instead of having to select from several hundreds of \xbrl tags, the expert would only have to inspect a short list of $k$ proposed tags.

\begin{figure}[h!] 
\centering

\includegraphics[width=0.9\columnwidth]{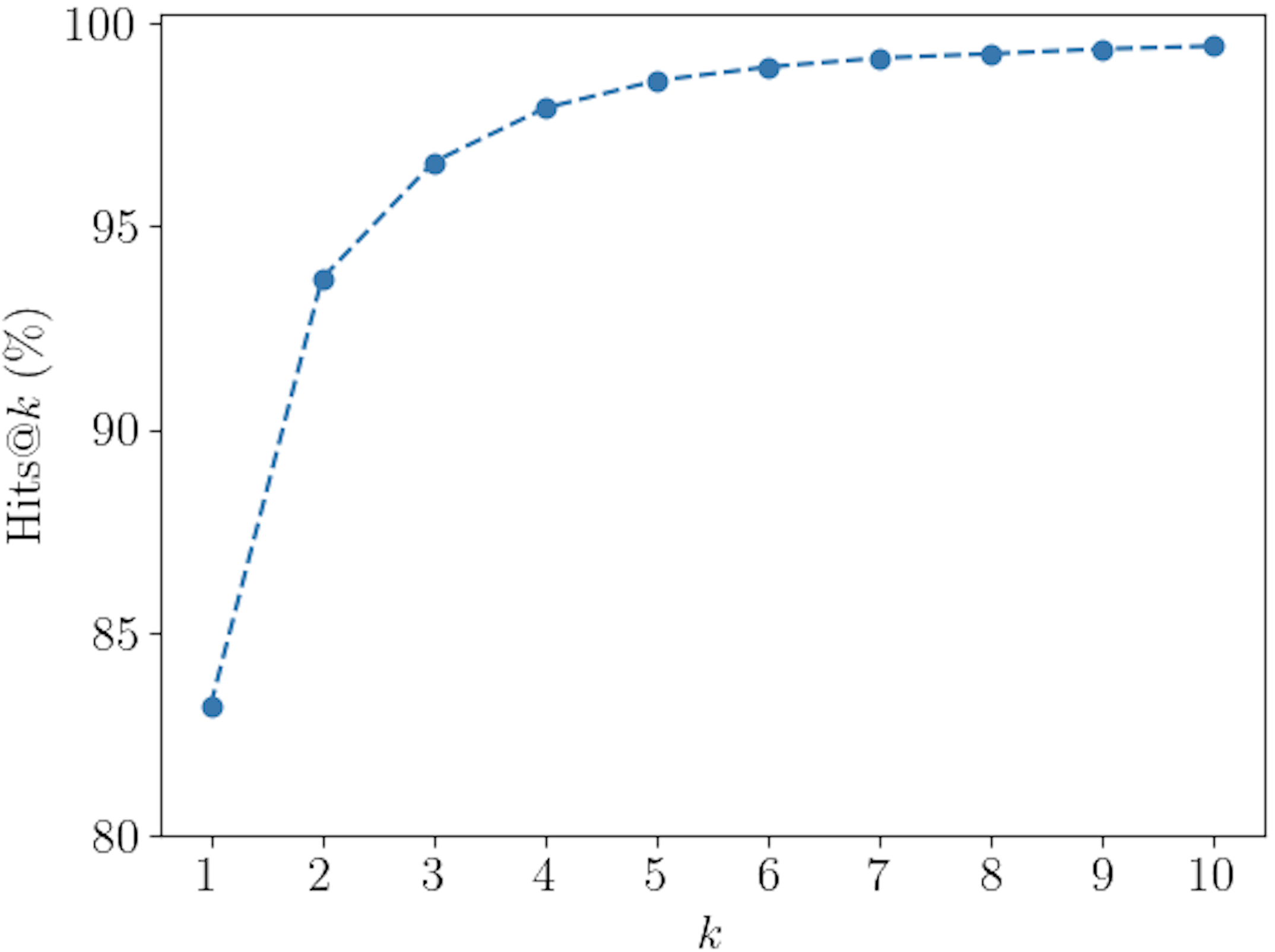}
\caption{$\mathrm{Hits@}k$ results (\%, avg.\ of 3 runs with different random seeds) on test data, for different $k$ values. Standard deviations were very small and are omitted.}
\label{fig:business_study}

\end{figure}

We evaluate our best model, \secbertmag, in this use case using $\mathrm{Hits@}k$. We use the model to return the $k$ most probable \xbrl tags for each token that needs to be annotated, now assuming that the tokens to be annotated are known. If the correct tag is among the top $k$, we increase the number of hits by one. Finally, we divide by the number of tokens to be annotated. Figure~\ref{fig:business_study} shows the results for different values of $k$. The curve is steep for $k=1$ to $5$ and saturates as $k$ approaches 10, where $\mathrm{Hits@}k$ is nearly perfect (99.4\%). In practice, this means that a user would have to inspect 10 recommended \xbrl tags instead of hundreds for each token to be annotated; and in most cases, the correct tag would be among the top 5 recommended ones.

\subsection{Error Analysis}\label{sec:analysis}
We also performed an exploratory data and error analysis to unveil the peculiarities of \finerdata, extract new insights about it, and discover the limitations of our best model. Specifically, we manually inspected the errors of \secbertmag in under-performing classes (where $\mathrm{F_1}<50\%$) and identified three main sources of errors.\vspace{2mm}

\noindent\textbf{Specialized terminology:} In this type of errors, the model is able to understand the general financial semantics, but does not fully comprehend highly technical details. For example, \textit{Operating Lease Expense} amounts are sometimes missclassified as \textit{Lease And Rental Expense}, i.e., the model manages to predict that these amounts are about expenses in general, but fails to identify the specific details that distinguish operating lease expenses from lease and rental expenses. Similarly, \textit{Payments to Acquire Businesses (Net of Cash Acquired)} amounts are mostly misclassified as \textit{Payments to Acquire Businesses (Gross)}. In this case, the model understands the notion of business acquisition, but fails to differentiate between net and gross payments.\vspace{2mm}

\noindent\textbf{Financial dates:} Another interesting error type is the misclassification of financial dates. For example, tokens of the class \textit{Debt Instrument Maturity Date} are mostly missclassified as not belonging to any entity at all (\textit{`O'} tag). Given the previous type of errors, one would expect the model to missclassify these tokens as a different type of financial date, but this is not the case here. We suspect that errors of this type may be due to annotation inconsistencies by the financial experts.\vspace{2mm}

\begin{figure}[t!] 
\centering

\fbox{
\includegraphics[width=0.95\columnwidth]{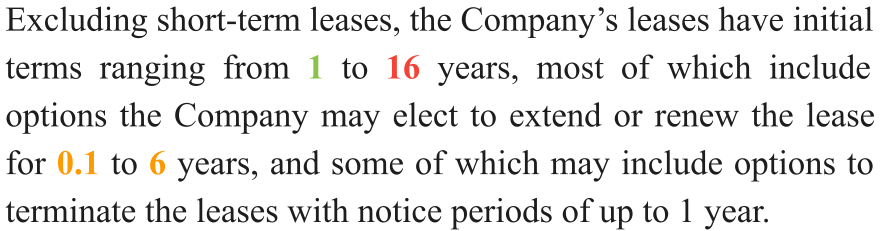}
}
\vspace*{-4mm}
\caption{A manually inspected sentence from \finerdata showing some inconsistencies in the gold \xbrl tags of the auditors. The green `1' is correctly annotated with the \xbrl tag \textit{Lessee Operating Lease Term Of Contract}. The red `16' should have also been annotated with the same tag, but is not, possibly because the annotator thought the (same) tag was obvious. The orange numbers `0.1' and `6' lack \xbrl annotations; they should have both been annotated as \textit{Lessee Operating Lease Renewal Term}. We can only speculate that the auditor might not have been aware that there is an \xbrl tag for lease renewal terms, in which case the recommendation engine of Section~\ref{sec:business_study} might have helped.}
\label{fig:inconsistencies}
\vspace*{-3mm}
\end{figure}

\noindent\textbf{Annotation inconsistencies:} 
Even though the gold \xbrl tags of \finerdata come from professional auditors, as required by the Securities \& Exchange Commission (\ussec) legislation, there are still some discrepancies. We provide an illustrative example in  Figure \ref{fig:inconsistencies}. We believe that such inconsistencies are inevitable to occur and they are a part of the real-world nature of the problem.\vspace{2mm}

\noindent We hope that this analysis inspires future work on \xbrl tagging. For example, the specialized terminology and financial date errors may be alleviated by adopting hierarchical classifiers \cite{lmtc1-chalkidis, structured1-manginas}, which would first detect entities in coarse classes (e.g., expenses, dates) and would then try to classify the identified entities into finer classes (e.g., lease vs.\ rent expenses, instrument maturity dates vs.\ other types of dates). It would also be interesting to 
train classifiers towards detecting wrong (or missing) gold annotations, in order to help in quality assurance checks of \xbrl-tagged documents. 

\section{Conclusions and Future Work}\label{sec:conclusions}
We introduced a new real-word \nlp task from the financial domain, \xbrl tagging, required by regulatory commissions worldwide. We released \finerdata, a dataset of 1.1M sentences with \xbrl tags. Unlike typical entity extraction tasks, \finerdata uses a much larger label set (139 tags), most tokens to be tagged are numeric, and the correct tag depends mostly on context rather than the tagged token. We experimented with several neural classifiers, showing that a \bilstm outperforms \bert due to the excessive numeric token fragmentation of the latter. We proposed two simple and effective solutions that use special tokens to generalize over the shapes and magnitudes of numeric expressions. We also experimented with \finbert, an existing \bert model for the financial domain, which also benefits from our special tokens. Finally, we pre-trained and released our own domain-specific \bert model, \secbert, both with and without the special tokens, which achieves the best overall results with the special tokens, without costly \crf layers.

In future work, one could hire experts to re-annotate a subset of the dataset to measure human performance against the gold tags. Future work could also consider less frequent \xbrl tags (few- and zero-shot learning) and exploit the hierarchical dependencies of \xbrl tags, possibly with hierarchical classifiers, building upon our error analysis.

\bibliographystyle{acl_natbib}
\bibliography{finer_acl_2022}
\newpage
\appendix

\section{Experimental Setup}\label{experimental-setup}

For \spacy, we followed the recommended practices.\footnote{\url{https://spacy.io/usage/v2-3}.} All other methods were implemented in \textmc{tensorflow}.\footnote{\url{https://www.tensorflow.org/}} 
Concerning \bert models, we used the implementation of \textmc{huggingface} \citep{Wolf2019HuggingFacesTS}.
We also use Adam \citep{adam-optimizer}, Glorot initialization \citep{glorot2010}, and the categorical cross-entropy loss.

Hyper-parameters were tuned on development data with Bayesian Optimization \cite{bayesian-optimization} monitoring the development loss for 15 trials.\footnote{We used \textmc{keras} \textmc{tuner} (\url{https://keras-team.github.io/keras-tuner/documentation/tuners/})} For the \bilstm encoders, we searched for $\{1,2,3\}$ hidden layers, $\{128, 200, 256\}$ hidden units, $\{1e$-$3, 2e$-$3, 3e$-$3, 4e$-$3, 5e$-$3\}$ learning rate, and $\{0.1, 0.2, 0.3\}$ dropout. We trained for $30$ epochs using early stopping with patience $4$. For \bert, we used grid-search to select the optimal learning rate from $\{1e$-$5, 2e$-$5, 3e$-$5, 4e$-$5, 5e$-$5\}$, fine-tuning for $10$ epochs, using early stopping with patience $2$. All final hyper-parameters are shown in Table \ref{appendix-exp}. 
Training was performed mainly on a \textmc{dgx} station with 4 \textmc{nvidia} \textmc{v}100 \textmc{gpu}s and an Intel Xeon \textmc{cpu} \textmc{e}5-2698 v4 @ 2.20\textmc{gh}z.

\begin{table}[h]
\centering
\resizebox{\columnwidth}{!}
{
\begin{tabular}{l|rccccc}
 \hline
 & $\mathit{Params}$ & $L$ & $U$ & $P_\mathrm{drop}$ & $\mathit{LR}$ \\
 \hline
 \bilstm (words) & 21M & 2 & 128 & 0.1 & 1e-3 \\
 \bilstm (subwords) & 8M & 1 & 256 & 0.2 & 1e-3 \\
 \bilstm (words) + \crf & 21M & 2 & 128 & 0.1 & 1e-3\\
 \bilstm (subwords) + \crf & 8M & 1 & 256 & 0.2 & 1e-3 \\
 \bilstmnum (subwords) & 1M & 1 & 256 & 0.2 & 1e-3\\
 \bilstmmag (subwords) & 0.8M & 2 & 128 & 0.1 & 1e-3\\
 \hline
 \bert & 110M  & - & - & - & 1e-5 \\
 \bert + \num & 110M  & - & - & - & 1e-5 \\
 \bert + \magn & 110M  & - & - & - & 1e-5 \\
 \bert + \crf & 110M  & - & - & - & 1e-5\\
 \hline
 \finbert & 110M & - & - & - & 2e-5 \\
 \finbert + \num & 110M & - & - & - & 2e-5 \\
 \finbert + \magn & 110M & - & - & - & 2e-5 \\
 \hline
 \secbert & 110M & - & - & - & 1e-5 \\
 \secbertnum & 110M & - & - & - & 1e-5 \\
 \secbertmag & 110M  & - & - & - & 1e-5 \\
 \hline 
 \hline
 
\end{tabular}
}
\caption{Number of total parameters ($\mathit{Params}$) and the best hyper-parameter values for each method; i.e., number of recurrent layers ($L$), number of recurrent units ($U$), dropout probability $P_\mathrm{drop}$, learning rate ($\mathit{LR}$).
}
\label{appendix-exp}
\end{table}

\section{Additional Results}\label{app:additional_results}
\vspace*{-1mm}
Table \ref{tab_app:baseline-results} shows micro-averaged Precision, Recall, and F1 for the development and test data, using all baseline methods. The macro-averaged scores are similar and we omit them for brevity.
Using a logistic regression (\lr) classification layer, \bilstm (words) surpasses \bert both in Precision and F1 score. However, when using a \crf layer on top, \bert outperforms \bilstm (words) in all measures. 

Table~\ref{tab_app:ablation-results} shows the micro-averaged Precision, Recall, and $\mathrm{F_1}$ for the development and test data using the \bilstm models operating on subwords with the proposed tokenizations. \num and \magn tokens help the model to bypass the word fragmentation problem, increasing its scores in all metrics.

\begin{table*}[ht]
\resizebox{\textwidth}{!}
{
\begin{tabular}{ l|ccc|ccc}
 \hline
 &
 \multicolumn{3}{c}{\textmc{development}} &  \multicolumn{3}{c}{\textmc{test}} \\
 & \microp & \micror & \microf   & \microp & \micror & \microf \\
 \hline

 \spacy
 & 38.2 $\pm$ 0.4
 & 58.2 $\pm$ 0.8 
 & 46.1 $\pm$ 0.1 
 & 40.8 $\pm$ 0.8 
 & 60.0 $\pm$ 0.4 
 & 48.6 $\pm$ 0.4 \\
 
 \bilstm (words)
 & 78.6 $\pm$ 2.4 
 & 80.3 $\pm$ 1.2
 & 79.4 $\pm$ 1.0
 & 75.4 $\pm$ 1.9
 & 78.0 $\pm$ 0.8 
 & 77.3 $\pm$ 0.6
 \\
 
 \bilstm (subwords)
 & 73.4 \stdsign 0.1
 & 77.2 \stdsign 0.0
 & 75.2 \stdsign 0.1
 & 68.8 \stdsign 0.2
 & 74.1 \stdsign 0.2
 & 71.3 \stdsign 0.2
 \\
 \bert (subwords)
 &  74.9 $\pm$ 1.5
& 82.0 $\pm$ 1.3
& 78.2$ \pm$ 1.4 
& 71.5 $\pm$ 1.1
& 79.6 $\pm$ 1.4
& 75.1 $\pm$ 1.1 \\
 \hline 
 \bilstm (words) + \crf 
 & 73.4 $\pm$ 2.0     
 & 69.3 $\pm$ 0.9    
 & 71.3 $\pm$ 1.2    
 & 70.9 $\pm$ 1.8     
 & 68.0 $\pm$ 0.9
 & 69.4 $\pm$ 1.2 \\
 
 \bilstm (subwords) + \crf 
 & \textbf{80.0} \stdsign 0.3
 & 78.7 \stdsign 0.5
 & 79.3 \stdsign 0.4
 & \textbf{76.5} \stdsign 0.2
 & 76.0 \stdsign 0.2
 & 76.2 \stdsign 0.2
 \\
 
 \bert (subwords) + \crf
 & 78.3 $\pm$ 0.8
 &\textbf{83.6} $\pm$ 0.4
 &\textbf{80.9} $\pm$ 0.3
 
 & 75.0 $\pm$ 0.9
 &\textbf{81.2} $\pm$ 0.2
 &\textbf{78.0} $\pm$ 0.5
 \\

 \hline
 
\end{tabular}
}
\caption{Entity-level micro-averaged P, R, F1 $\pm$ std.\ dev.\ (3 runs) on the dev.\ and test data for our 
baselines.}
\label{tab_app:baseline-results}

\vspace{5mm} 
\centering
\resizebox{\textwidth}{!}
{
\begin{tabular}{ l|ccc|ccc}
 \hline
 &
 \multicolumn{3}{c}{\textmc{development}} &  \multicolumn{3}{c}{\textmc{test}} \\
 & \microp & \micror & \microf   & \microp & \micror & \microf \\
 \hline
 \bilstm (subwords)
 & 73.4 \stdsign 0.1
 & 77.2 \stdsign 0.0
 & 75.2 \stdsign 0.1
 & 68.8 \stdsign 0.2
 & 74.1 \stdsign 0.2
 & 71.3 \stdsign 0.2
 \\
 \bilstm (subwords) + \crf 
 & 80.0 \stdsign 0.3
 & 78.7 \stdsign 0.5
 & 79.3 \stdsign 0.4
 & 76.5 \stdsign 0.2
 & 76.0 \stdsign 0.2
 & 76.2 \stdsign 0.2
 \\
 \hline
 \bilstmnum (subwords)
 & 77.9 \stdsign 0.4
 & 78.6 \stdsign 0.7
 & 78.2 \stdsign 0.6
 & 74.8 \stdsign 0.2
 & 76.5 \stdsign 0.5
 & 75.6 \stdsign 0.3
 \\
 \bilstmmag (subwords)
 & \textbf{81.1} \stdsign 0.1
 & \textbf{81.5} \stdsign 0.3
 & \textbf{81.3} \stdsign 0.2
 & \textbf{77.5} \stdsign 0.3
 & \textbf{78.7} \stdsign 0.5
 & \textbf{78.1} \stdsign 0.4
 \\
 
 \hline
 
\end{tabular}
}
\caption{Entity-level micro-averaged P, R, F1 $\pm$ std.\ dev.\ (3 runs) on the dev.\ and test data for the \bilstm models using the \num and \magn tokens.}
\label{tab_app:ablation-results}
\vspace{14cm}
\end{table*}

\end{document}